\documentclass[11pt]{article}
\usepackage{amsmath}
\usepackage[utf8]{inputenc}
\usepackage{amssymb}
\usepackage{xspace}

\usepackage[preprint]{acl}

\usepackage{times}
\usepackage{latexsym}
\usepackage{tabularx}
\usepackage{booktabs}

\usepackage{listings}
\usepackage{xcolor}
\usepackage{fvextra}
\usepackage{booktabs}
\usepackage{array}
\usepackage{multirow}
\usepackage{float}
\usepackage{placeins}
\usepackage{fancyvrb}
\usepackage[most]{tcolorbox}

\definecolor{bugred}{RGB}{220,50,50}
\definecolor{conceptred}{RGB}{180,0,0}

\usepackage[T1]{fontenc}

\usepackage[utf8]{inputenc}

\usepackage{microtype}
\newcommand{\instructkg}{\textsc{InstructKG}\xspace}

\newcommand\blfootnote[1]{%
  \begingroup
  \renewcommand\thefootnote{}\footnote{#1}%
  \addtocounter{footnote}{-1}%
  \endgroup
}

\usepackage{inconsolata}

\usepackage{graphicx}

\hypersetup{
  pdftitle={Instructor-Aligned Knowledge Graphs for Personalized Learning},
  pdfauthor={Abdulrahman AlRabah, Priyanka Kargupta, Jiawei Han, Abdussalam Alawini},
  pdfkeywords={knowledge graph construction, educational NLP, concept extraction}
}

\title{Instructor-Aligned Knowledge Graphs for Personalized Learning}

\author{Abdulrahman AlRabah, Priyanka Kargupta, Jiawei Han, Abdussalam Alawini \\
  University of Illinois Urbana-Champaign \\
  Siebel School of Computing and Data Science \\
  Urbana, IL, USA \\
  \texttt{\{alrabah2,pk36,hanj,alawini\}@illinois.edu}}

\begin{document}
\maketitle
\begin{abstract}
Mastering educational concepts requires understanding both their prerequisites (e.g., recursion before merge sort) and sub-concepts (e.g., merge sort as part of sorting algorithms). Capturing these dependencies is critical for identifying students' knowledge gaps and enabling targeted intervention for personalized learning, especially in large-scale courses where instructors cannot feasibly diagnose individual misunderstanding. While knowledge graphs offer a natural representation for capturing these conceptual relationships at scale, existing approaches are either surface-level or disregard the rich pedagogical signals embedded in instructional materials. We propose \instructkg, a framework for automatically constructing instructor-aligned knowledge graphs that capture a course's intended learning progression. Given a course's lecture materials, \instructkg extracts significant concepts as nodes and infers learning dependencies as directed edges (e.g., ``part-of'' or ``depends-on'' relationships). The framework synergizes the rich temporal and semantic signals unique to educational materials (e.g., ``recursion'' is taught before ``mergesort'') with the generalizability of large language models. Through experiments on diverse lecture materials, human evaluation, and a prerequisite-learning benchmark, we demonstrate that \instructkg captures rich, instructor-aligned learning progressions.
\end{abstract}

\blfootnote{Code: \url{https://github.com/aalrabah/instructkg}}

\vspace{-5mm}
\section{Introduction}
Learning educational concepts is incremental, with each new idea building on previously acquired foundations. Understanding these conceptual dependencies, such as knowing that recursion must be understood before merge sort, or that merge sort is a specific instance of sorting algorithms, is critical for effective learning. When students lack awareness of these prerequisite relationships, they often struggle to identify knowledge gaps, leading to compounding difficulties as courses progress \cite{simonsmeier2022domain, valstar2019relationship}.

This challenge is particularly acute in large-scale educational settings, where instructors cannot feasibly diagnose individual misconceptions or provide personalized guidance on which foundational concepts require reinforcement \cite{chowdhury2025large}.

\begin{figure*}[!t]
  \centering
  \includegraphics[width=0.85\textwidth]{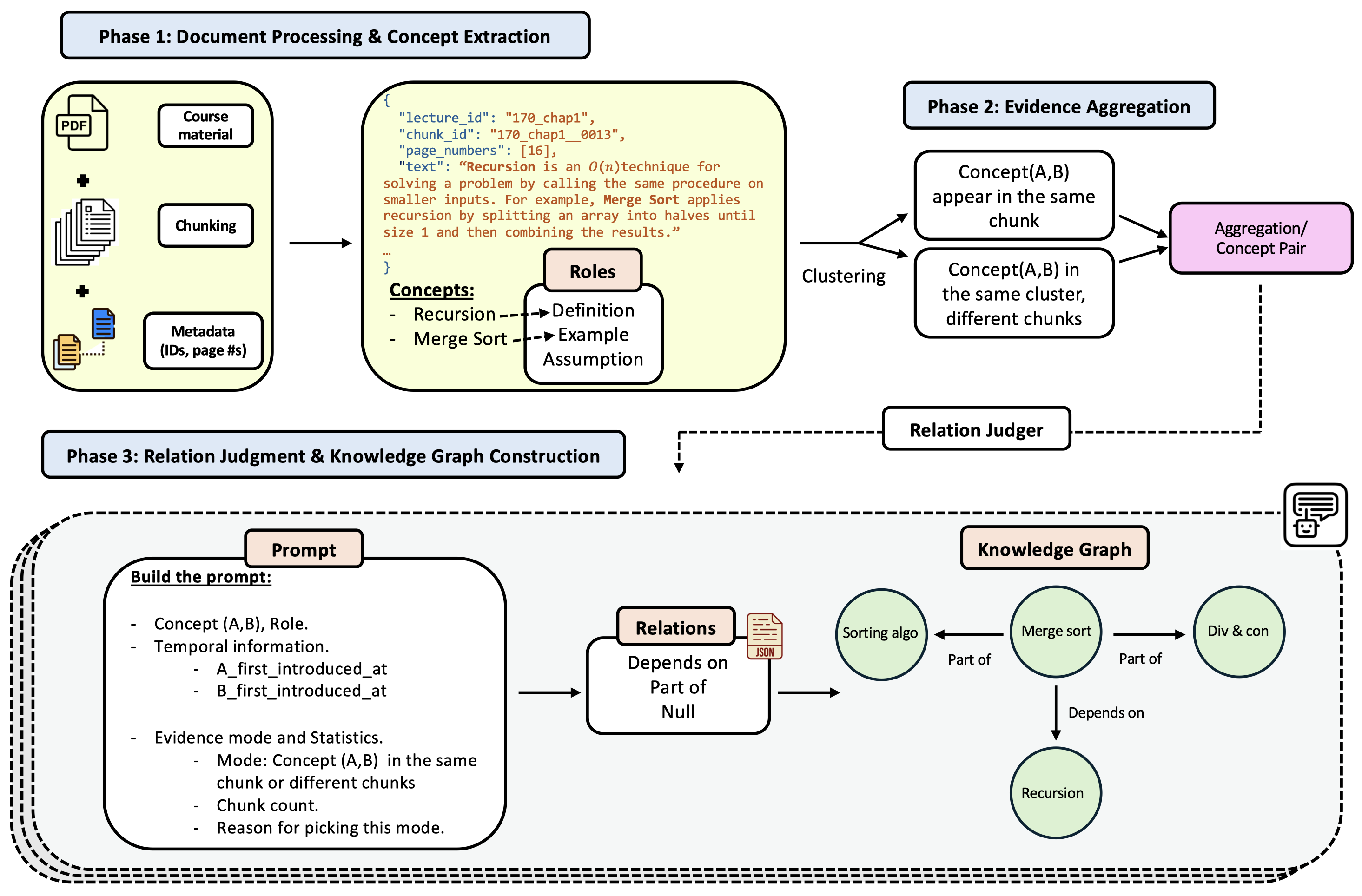}
  \caption{Overview of the \instructkg framework showing the three-phase pipeline.}
  \label{fig:system}
\end{figure*}

As illustrated in Figure~\ref{fig:system}, a lecture introducing merge sort may assume recursion from earlier instruction, and a student who missed recursion then struggles downstream. Capturing such implicit prerequisite and compositional dependencies lets automated systems pinpoint where understanding breaks down and target remediation.

Knowledge graphs naturally encode such dependencies at scale, but existing approaches miss the fine-grained, instructor-aligned relationships in course materials. Knowledge-base and metadata-driven methods identify prerequisites via co-occurrence or the Semantic Web~\cite{chanaa2024prerequisites, ain2023automatic}, but operate over broad topics rather than course-specific concepts~\cite{wang2024multi, zhao2022edukg}. Text-based extraction, from OpenIE~\cite{yamamoto2025exploring} to recent LLM pipelines with multi-stage canonicalization~\cite{mo2025kggen, zhang2024extract, wang2025adaptive, huang2025can, zhang2025gkg}, processes each passage independently, ignoring instructional signals like teaching order and pedagogical role. Our experiments confirm that even strong LLM-based extraction, without instructor-grounded signals, falls well short of the full method, particularly in relationship coverage.

Instructional materials contain rich, underutilized signals for inferring dependencies: \emph{temporal} signals (teaching order, where earlier concepts serve as prerequisites) and \emph{semantic} signals (pedagogical role, where a concept in another's definition suggests a foundational dependency). Both are unique to educational content and overlooked by current methods.

We propose \instructkg, a framework for automatically constructing instructor-aligned knowledge graphs from lecture materials (slides, notes, transcripts). \instructkg extracts concepts as nodes and infers directed dependency edges of two types, depends-on (prerequisite) and part-of (compositional), by combining the temporal and semantic signals in educational materials with LLM reasoning, without external knowledge bases or manual annotation.

\noindent\textbf{Our contributions are as follows:}
\begin{itemize}
  \item We introduce a methodology that leverages temporal signals (teaching order) and semantic signals (concept roles in context) unique to instructional materials to ground LLM-based reasoning in instructor-aligned evidence, rather than relying on the model's parametric knowledge alone. 
  \item We design a cluster-based evidence mechanism that identifies potential concept relationships even when concepts do not co-occur within the same instructional segment, which captures implicit dependencies beyond chunk-level co-occurrence.
  \item We conduct experiments on real-world lecture materials across multiple courses and perform human evaluation, demonstrating that \textsc{InstructKG} captures rich learning progressions aligned with pedagogical intent. 
\end{itemize}

\section{Related Works}
\noindent\textbf{Knowledge Graph Construction.} Methods for constructing knowledge graphs from text range from rule-based and OpenIE approaches to supervised neural extraction and, more recently, LLM-based prompting~\cite{mo2025kggen, zhang2024extract, kolluru2020openie6, wadden2019entity, wang2023instructuie, jiao2023instruct}. Supervised neural systems jointly extract entities and relations via span-based, graph-based, or marker-based formulations~\cite{eberts2019span, lin2020joint, zhong2021frustratingly, wang2020tplinker, ye2022packed}. Recent LLM pipelines extract triples and canonicalize them through multi-stage processing: KGGen~\cite{mo2025kggen} merges duplicates via iterative clustering, EDC~\cite{zhang2024extract} uses an extract–define–canonicalize pipeline with embedding retrieval and LLM verification, and others add fine-tuned graph judgment~\cite{huang2025can}, schema-constrained normalization~\cite{wang2025adaptive}, curriculum learning~\cite{zhang2025gkg}, and OpenIE with LLM validation~\cite{yamamoto2025exploring}.

However, these methods are designed for general-purpose knowledge graph construction and do not leverage signals specific to educational materials. None incorporate teaching order, pedagogical role classification, or cross-segment evidence aggregation to infer concept dependencies.

\noindent\textbf{Educational Knowledge Graphs and Prerequisite Learning.} Prior work retrieves prerequisite candidates via SPARQL over DBpedia and Wikidata~\cite{chanaa2024prerequisites,ain2023automatic,wang2025adaptive}, frames prerequisite detection as co-occurrence-based binary classification~\cite{chanaa2024prerequisites}, or extracts relations from textbooks via ontology-guided NER and OpenIE~\cite{zhao2022edukg}. These methods are limited: external knowledge bases miss instructor-specific definitions and course-level nuance~\cite{zhao2022edukg, wang2025adaptive, aytekin2024ace}, co-occurrence signals are too coarse for fine-grained prerequisites~\cite{chanaa2024prerequisites}, and ontology-guided methods need substantial annotation~\cite{zhao2022edukg}. In contrast, \instructkg extracts dependencies directly from lecture materials using temporal and semantic signals to build instructor-aligned graphs. 

A complementary line learns concept prerequisite relations against human annotated labels, including \citet{pan2017prerequisite} on MOOC transcripts, LectureBank \citep{li2019should}, and AL-CPL \citep{liang2018investigating}. These methods train supervised classifiers on the labels over fixed concept vocabularies, whereas \instructkg{} builds course local graphs without label supervision. We evaluate \instructkg{} directly on the MOOC benchmark in Section~\ref{sec:mooc-benchmark} and use LectureBank to validate our LLM judge in Section~\ref{sec:eval_met}.

\noindent\textbf{Concept Extraction from Educational Content.} Prior methods extract concepts from educational materials via unsupervised embedding-based clustering~\cite{huang2023werece} or distantly supervised NER~\cite{lu2023distantly}, but identify concept mentions without inferring relationships. Others model sequential structure at the lecture level, clustering MOOC lectures into precedence graphs~\cite{alsaad2020unsupervised} or learning topic transitions~\cite{alsaad2021topic}, but operate on lectures rather than concepts. In contrast, \instructkg operates at the concept level, leveraging pedagogical roles and cross-segment evidence to infer fine-grained dependencies directly from lecture materials.

\noindent\textbf{LLMs for Information Extraction.} Recent work uses LLMs for KG construction: GenIE~\cite{josifoski2022genie} autoregressively generates schema-grounded triplets, and CodeKGC~\cite{bi2024codekgc} frames triple extraction as code generation with schema-aware prompts. These target KB-grounded factual triplets constrained by predefined schemas; they model neither pedagogical relations (prerequisites, part-of) nor educational signals like concept roles or teaching order, which \instructkg leverages to infer instructor-aligned dependencies.

\section{Methodology}
We propose \instructkg, a framework for automatically constructing instructor-aligned knowledge graphs from course materials by extracting fine-grained concepts and inferring two learning dependencies between them: 
\texttt{depends\_on} (prerequisite) and \texttt{part\_of} (compositional). Its core idea is to condition LLM-based relation inference on three pedagogical signals absent from general-purpose KG construction: (i) the \emph{temporal order} in which concepts are introduced, (ii) the \emph{pedagogical role} a concept plays in each chunk (Definition, Example, Assumption), and (iii) \emph{cross-chunk semantic evidence} linking concepts separated by chunking boundaries. We operationalize this in three stages, shown in Figure~\ref{fig:system}: concept and role extraction, evidence aggregation, and relation judgment.

\subsection{Problem Formulation}

Given an ordered set of lecture documents $\mathrm{D}=\{\mathrm{d}_1,\ldots,\mathrm{d}_n\}$ (PDFs of slides, notes, or other material, sorted by teaching sequence), our goal is a directed knowledge graph $G=(V,E)$ with $E \subseteq V \times V \times R$ and $R=\{\texttt{depends\_on},\texttt{part\_of}\}$: an edge $(A,B,\texttt{depends\_on})$ means $A$ requires $B$ as a prerequisite (e.g., merge sort \texttt{depends\_on} recursion), and $(A,B,\texttt{part\_of})$ means $A$ is a component or subtype of $B$ (e.g., merge sort \texttt{part\_of} sorting algorithms). For each occurrence of a concept $v$ in a chunk $c$ we assign a role $r(v,c)\in\{\textsc{Definition},\textsc{Example},\textsc{Assumption},\textsc{NA}\}$: being introduced, demonstrated, referenced as prior knowledge, or none of these.

\noindent\textbf{Pre-processing.} We sort PDFs into teaching order by lecture or chapter numbers in filenames (e.g., ``lecture-03''), then split each into chunks under a maximum token limit, merging adjacent related segments. This yields a corpus $\mathcal{C}=\{c_1,\ldots,c_m\}$ where each chunk carries its lecture identifier, chunk index, and page numbers as \texttt{lecture\_id\_\_chunk\_index}.

\noindent\textbf{Concept Extraction.} For each chunk $c \in \mathcal{C}$, we prompt an open-source, instruction-tuned LLM (specific models detailed in Section~\ref{sec:exp-setup}) to extract meaningful course concepts, excluding code tokens, variable names, and example values. Extracted concepts are deduplicated case-insensitively while preserving order. Each concept is assigned a canonical identifier via deterministic normalization: the concept string is uppercased and non-alphanumeric characters are replaced with underscores (e.g., ``Left outer join'' $\rightarrow$ \texttt{LEFT\_OUTER\_JOIN}).

\subsection{Mapping Pedagogical Roles to Concepts}
How a concept is used signals its relationships: a concept appearing as an \textsc{Assumption} where another is \textsc{Defined} signals a prerequisite (e.g., \textit{merge sort} \texttt{depends\_on} \textit{recursion}), while an \textsc{Example} alongside another's \textsc{Definition} suggests a compositional link (e.g., \textit{merge sort} \texttt{part\_of} \textit{sorting algorithms}). We use \textsc{Definition}, \textsc{Example}, and \textsc{Assumption} as the most informative roles; others (e.g., Application, Comparison) are excluded.

\noindent\textbf{Role Classification.} For each (chunk, concept) pair, we classify the concept's role into the four categories defined above. The output is a set of \emph{mention} records $\mathcal{M}$, where each mention $m=$ (\texttt{concept\_id}, \texttt{chunk\_id}, \texttt{lecture\_id}, \texttt{role}) captures a concept's occurrence and pedagogical context. We also track the first introduction of each concept by teaching order:
\begin{equation}
\operatorname{first}(v)=\!\!argmin_{m \in \mathcal{M}(v)}\!\bigl(\ell(m),\, c(m)\bigr),
\end{equation}
where $\ell(m)$ and $c(m)$ are the lecture and chunk indices of mention $m$.

\subsection{Context Clustering}

Chunk-level co-occurrence captures concepts related within the same chunk, but pedagogically related concepts are often separated by chunking boundaries, for instance sorting algorithms introduced at the start of a chapter while merge sort is defined on the next page. To surface these implicit relationships across boundaries, we cluster semantically similar chunks. We embed each chunk using a sentence-transformer model to obtain vector representations $\mathbf{x}_c \in \mathbb{R}^d$ for each chunk $c$. We then apply UMAP for dimensionality reduction (using cosine distance) followed by HDBSCAN for density-based clustering. This assigns each chunk a cluster label $\ell_c \in \{-1,0,1, \ldots, k\}$, where $-1$ denotes noise.

For each concept mapped to a cluster, we select representative chunks by computing the cluster centroid $\boldsymbol{\mu}=\frac{1}{|C|} \sum_{c \in C} \mathbf{x}_c$ and ranking chunks by cosine similarity to $\boldsymbol{\mu}$. These representative chunks serve as the most relevant evidence provided to the LLM during relation judgment. Finally, we enrich each cluster with its concept set by building a mapping from chunk identifiers to concepts and taking the union over all chunks in the cluster. Together with chunk-level co-occurrence, this allows us to identify which concepts are likely to have a direct relationship from both a local (chunk) and global (corpus) level.

\noindent\textbf{Temporal \& Role-Grounded Evidence.}
For each candidate pair $(A,B)$, we build an evidence packet aggregating the prior phases' signals: each concept's pedagogical roles, temporal information on where it was first introduced, and the relevant evidence text. Candidates come from two sources, chunk co-occurrence (same chunk) and cluster co-occurrence (same thematic cluster, different chunks); pairs with neither are excluded. Each packet is then passed to the relation judgment phase.
\vspace{-6mm}
\subsection{Relation Judgment and Knowledge Graph Construction}
The final phase uses an LLM to determine the relation type for each candidate pair from the aggregated evidence. We select evidence by priority: if the pair co-occurs in a chunk, we provide chunks containing both concepts (the most direct evidence); otherwise we provide separate chunks for $A$ and $B$ from their shared cluster; pairs with neither are skipped. We then prompt the LLM with the pair, their roles, temporal information, and selected chunks to decide whether $A$ \texttt{depends\_on} $B$ (prerequisite), $A$ is \texttt{part\_of} $B$ (component/subtype), or no relationship holds, always with direction $A\!\to\!B$ and a justification grounded in the evidence; pairs are normalized alphabetically so $(A,B)$ and $(B,A)$ map to one candidate (full prompts in Appendix~\ref{app:prompts}). The final graph $\mathcal{G}=(\mathcal{V},\mathcal{E})$ collects all non-null judgments as directed edges, each carrying its evidence and justification for interpretability.

\section{Experimental Design}

\begin{table}[htbp]
\centering
\small
\caption{Dataset statistics for the five evaluated courses.}
\label{tab:datasets}
\resizebox{\columnwidth}{!}{%
\begin{tabular}{lcrrr}
\toprule
\textbf{Course} & \textbf{Format} & \textbf{Lectures} & \textbf{Pages} & \textbf{Chunks} \\
\midrule
Database Systems & Slides        & 27 & 569 & 552 \\
NLP              & Slides + Notes & 15 & 971 & 962 \\
Algorithms       & Lecture Notes  &  9 & 222 & 221 \\
Finance          & Lecture Notes  &  7 & 530 & 478 \\
Mech. Eng.       & Book Chapters  &  2 & 120 & 125 \\
\bottomrule
\end{tabular}%
}
\end{table}

\begin{table*}[t]
\centering
\small
\setlength{\tabcolsep}{5pt}
\renewcommand{\arraystretch}{1.15}
\caption{Per-course node- and triplet-level significance (\%), judged by an LLM evaluator. Context only removes clustering, role, and temporal signals, keeping the same concept extraction. For each course, the best Node and best Triplet values are in bold; ties are both bolded.}
\label{tab:per-course-results}
\resizebox{\textwidth}{!}{%
\begin{tabular}{ll *{5}{cc}}
\toprule
\multirow{2}{*}{\textbf{Method}} & \multirow{2}{*}{\textbf{Model}}
& \multicolumn{2}{c}{\textbf{Algorithms}}
& \multicolumn{2}{c}{\textbf{NLP}}
& \multicolumn{2}{c}{\textbf{Database Systems}}
& \multicolumn{2}{c}{\textbf{Finance}}
& \multicolumn{2}{c}{\textbf{Mech. Eng.}} \\
\cmidrule(lr){3-4} \cmidrule(lr){5-6} \cmidrule(lr){7-8} \cmidrule(lr){9-10} \cmidrule(lr){11-12}
& & Node & Trip. & Node & Trip. & Node & Trip. & Node & Trip. & Node & Trip. \\
\midrule
\multirow{3}{*}{\shortstack[l]{EDC \citep{zhang2024extract}\\\scriptsize EMNLP 2024}}
& Llama-3B  & 44 & 20 & 45 & 13 & 48 & 11 & 50 & 13 & 64 & 23 \\
& Llama-8B  & 51 & 14 & 46 & 14 & 48 & 13 & 47 & 8  & 70 & 15 \\
& Qwen2.5-14B  & 53 & 17 & 53 & 18 & 53 & 16 & 52 & 11 & 75 & 17 \\
\midrule
\multirow{3}{*}{\shortstack[l]{KG-Gen \citep{mo2025kggen}\\\scriptsize NeurIPS 2025}}
& Llama-3B  & 44 & 25 & 50 & 35 & 53 & 31 & 59 & 30 & 70 & 26 \\
& Llama-8B  & 53 & 38 & 53 & 39 & 53 & 40 & 64 & 39 & 70 & 28 \\
& Qwen2.5-14B  & 51 & 41 & 54 & 48 & 58 & 43 & 63 & 39 & 76 & 44 \\
\midrule
\multirow{3}{*}{\instructkg}
& Llama-3B  & 94 & 29 & 86 & 36 & 88 & 36 & 93 & 28 & 79 & 9  \\
& Llama-8B  & 93 & 34 & 85 & 48 & 88 & 43 & 97 & 30 & 78 & 21 \\
& Qwen2.5-14B  & \textbf{98} & \textbf{52} & \textbf{97} & \textbf{63} & 95 & \textbf{61} & \textbf{98} & \textbf{58} & \textbf{87} & \textbf{85} \\
\midrule
\multirow{3}{*}{$\times$ Clustering}
& Llama-3B  & 93 & 37 & 85 & 40 & 87 & 38 & 93 & 32 & 79 & 18 \\
& Llama-8B  & 93 & 36 & 87 & 49 & 88 & 45 & 97 & 35 & 77 & 20 \\
& Qwen2.5-14B  & 97 & 46 & 96 & 59 & 94 & 60 & 95 & 17 & 83 & 23 \\
\midrule
\multirow{3}{*}{$\times$ Roles}
& Llama-3B  & 94 & 30 & 85 & 36 & 87 & 37 & 92 & 30 & 80 & 9  \\
& Llama-8B  & 93 & 35 & 85 & 47 & 87 & 43 & 97 & 36 & 78 & 19 \\
& Qwen2.5-14B  & 93 & 36 & 96 & 58 & 95 & 53 & 98 & 43 & 84 & 22 \\
\midrule
Context only 
& Qwen2.5-14B  & 97 & 47 & \textbf{97} & 61 & \textbf{97} & 60 & \textbf{98} & 55 & 86 & 66 \\
\bottomrule
\end{tabular}%
}
\end{table*}

\subsection{Datasets}
We evaluate \instructkg on five real-world courses from four universities spanning diverse domains (Table~\ref{tab:datasets}): Database Systems (relational databases, SQL, query optimization, transactions), Natural Language Processing (text processing, language models, neural approaches), Algorithms (algorithm design and analysis), Finance (valuation, risk, portfolios), and Mechanical Engineering (automotive technology). These vary in format, size, and instructional density, letting us assess \instructkg's robustness across teaching styles and course scales.

\noindent\textbf{Ablations.} To isolate each component we ablate \textbf{No Clustering} (chunk-level co-occurrence only), \textbf{No Roles} (judge sees evidence and temporal information but no role labels), and \textbf{Context only} (extraction unchanged, but clustering, role, and temporal signals all removed). No Clustering and No Roles run on all three models; Context only uses Qwen2.5-14B.

\subsection{Baselines}

We compare against two recent LLM-based KG construction methods, adapted to our schema. \textbf{KGGen} \cite{mo2025kggen} extracts triples and clusters entities and relations into canonical forms; \textbf{EDC} \cite{zhang2024extract} is an extract, define, canonicalize pipeline with embedding retrieval and LLM verification. Both run on the same chunked lecture input with DAG constraints enforced; full adaptation details are in Appendix~\ref{app:baselines}.

\subsection{Evaluation Metrics}
\label{sec:eval_met}
We use two excerpt-grounded metrics, each scored $\{0,1,2\}$ from retrieved excerpts and reported as a percentage. \emph{Node significance} measures whether a node is core course content (topics, methods, principles) rather than logistics or off-topic material. \emph{Triplet accuracy} measures whether a directed, typed edge (\texttt{depends\_on}, \texttt{part\_of}, \texttt{None}) has the correct relation and direction, with 2 requiring explicit excerpt support and partial or reversed relations scoring lower. Our primary evaluation uses expert human annotators (Section~\ref{sec:human-eval}). To scale, we add an LLM judge, \texttt{gpt-oss-120b} at temperature 0, constrained to structured JSON grounded on excerpts. We validate it against the full human evaluation sample, 100 nodes and 100 triplets per course across all five courses: it matches the human score exactly on 56\% of nodes and 38\% of triplets and fully contradicts on only 16\% and 17\%, scoring conservatively on triplets because it credits only what the excerpts state. All methods are graded identically, so we use the judge for system comparison and the human evaluation for absolute quality. As an external check on 41{,}921 human labeled LectureBank pairs, the judge reaches F1 0.711, within the 0.68 to 0.74 range of supervised methods trained on those labels \citep{li2019should}.

\subsection{Experimental Setup}\label{sec:exp-setup}
We evaluate \textsc{InstructKG} and baselines with three LLMs of varying scale: Llama-3.2-3B-Instruct, Llama-3.1-8B-Instruct \cite{dubey2024llama}, and Qwen2.5-14B \cite{qwen2.5}. Each model is used consistently across all \textsc{InstructKG} phases (concept extraction, role classification, relation judgment) and for KGGen and EDC, with temperature $0.1$ for deterministic outputs. For clustering, we embed chunks with \textit{all-MiniLM-L6-v2} (normalized), reduce with UMAP ($n_{\text{components}}{=}15$, $n_{\text{neighbors}}{=}15$), and cluster with HDBSCAN ($\texttt{min\_cluster\_size}{=}5$).  A representative course takes 55 minutes end to end on a single GPU with Qwen2.5-14B, dominated by relation judgment (48 minutes for 3{,}373 pairs); across all corpora runtimes range from 3 minutes to 3.6 hours (Appendix~\ref{app:runtime}). Main experiments use NVIDIA H200 GPUs; the representative-course runtime was measured on an RTX PRO 6000 class GPU; full hyperparameters are in Appendix~\ref{app:hyperparams}.

\section{Experimental Results}
We compare \instructkg against both baselines across all five courses and model scales. Results are evaluated on two dimensions: node-level significance, which measures whether extracted concepts are meaningful to the course, and triplet-level accuracy, which measures whether inferred relationships are pedagogically correct.

\subsection{Main Results}
\instructkg consistently outperforms both baselines across nearly all datasets and model scales (Table~\ref{tab:per-course-results}). At the node level the gains are substantial: with Qwen2.5-14B, \instructkg  reaches node significance of 98\%, 97\%, 95\%, and 98\% on Algorithms, NLP, Database Systems, and Finance, versus the strongest baseline (\textsc{KGGen}-14B) at 51\%, 54\%, 58\%, and 63\%, a relative improvement of over 70\% on average. Even at 3B scale, \instructkg (94\%, 86\%, 88\%, 93\%) roughly doubles both baselines, which stay in the 44--76\% range regardless of size, indicating that concept quality is driven by our instructor-grounded pipeline rather than model capacity alone.

For triplet accuracy, \instructkg with Qwen2.5-14B achieves 52\%, 63\%, and 61\% on Algorithms, NLP, and Database Systems, outperforming the best baseline (\textsc{KGGen}-14B: 41\%, 48\%, 43\%) by 27\%, 31\%, and 42\%. At smaller scales it is competitive with \textsc{KGGen} on the larger courses and consistently outperforms \textsc{EDC}, whose triplet accuracy stays below 25\% throughout, suggesting its canonicalization approach struggles to produce meaningful relationships even when it identifies relevant concepts. On the smallest course, Mechanical Engineering, \instructkg reaches 85\%, its largest margin over both baselines.

Across methods, performance improves with model size, with the largest gains between 8B and 14B, most pronounced for \instructkg's triplet accuracy (e.g., 34\% to 52\% on Algorithms, 43\% to 61\% on Database Systems), indicating that relation judgment benefits from stronger reasoning. Notably, these results hold across courses differing in format (slides, notes, book chapters), size (2 to 27 lectures), and domain, demonstrating \instructkg's robustness to varying instructional styles.

\noindent\textbf{Ablations.} At the node level all ablations stay within a few points of the full method (93\% or higher on the four larger courses), since these components affect relationship inference, not concept extraction. At the triplet level the full method leads on every course, and Context only explains why: on candidate pairs shared with it, per edge accuracy is essentially unchanged (mean difference 0.004), so role and temporal context contribute little direct accuracy at these scales. The advantage comes instead from clustering, which supplies roughly half of all candidate pairs and yields about twice as many correct edges at equal precision (466 vs.\ 236 on Finance). The two-lecture Mechanical Engineering course, with little cross-lecture structure to aggregate, is the exception.

\noindent\textbf{Error Analysis.} We examine the 1{,}591 triplets scored wrong or partial with Qwen2.5-14B. Only 1\% connect unrelated concepts; 49\% are plausible but unsupported by the retrieved evidence, 28\% reverse the direction, and 22\% assign the wrong relation type. The pipeline therefore rarely invents relationships, and its errors concentrate in direction and typing rather than in concept pairing. Evidence based filtering does not help: keeping only edges with two or more supporting chunks discards half the edges while leaving accuracy unchanged (0.56 vs.\ 0.57; full curve in Appendix~\ref{app:filter}). Instead, each edge carries its supporting excerpts and justification, which allows downstream tools to present the evidence rather than assert an unverified link.

\begin{table}[t]
\centering
\small
\setlength{\tabcolsep}{4pt}
\caption{The MOOC prerequisite benchmark \citep{pan2017prerequisite} and zero-shot \instructkg{} results (Qwen2.5-14B).}
\label{tab:mooc-benchmark}
\begin{tabularx}{\columnwidth}{@{}Xcc@{}}
\toprule
\textbf{Metric} & \textbf{ML} & \textbf{DSA} \\
\midrule
\multicolumn{3}{@{}l}{\textit{Benchmark statistics}} \\
Courses / videos & 2 / 548 & 3 / 449 \\
Labeled pairs & 6{,}712 & 2{,}540 \\
Positive pairs & 1{,}735 & 479 \\
\addlinespace
\multicolumn{3}{@{}l}{\textit{\instructkg{} (zero-shot)}} \\
Precision & 0.43 & 0.35 \\
Recall & 0.29 & 0.24 \\
F1 & 0.35 & 0.29 \\
\addlinespace
Direction accuracy & 0.63 & 0.62 \\
\quad + teaching-order correction & 0.62 & 0.66 \\
\bottomrule
\end{tabularx}
\end{table}

\begin{table*}[t]
\centering
\renewcommand{\arraystretch}{1.0}
\caption{SQL Question--Concept Mapping. Red indicates error-related concepts. Dependency order: \texttt{ORDER BY} $\rightarrow$ \texttt{WHERE} $\rightarrow$ \texttt{FROM}.}
\label{tab:question-concept-mapping}
\scriptsize
\setlength{\tabcolsep}{4pt}
\begin{tabular}{p{0.31\textwidth} p{0.31\textwidth} p{0.31\textwidth}}
\toprule
\textbf{Q1} & \textbf{Q2} & \textbf{Q3} \\
\midrule
\textbf{Question.} Resources \textbf{NOT} accessed by users in Segment A. Average duration per resource.
&
\textbf{Question.} List open tickets, sorted by most recently updated first.
&
\textbf{Question.} Total hours per project (must use time logs as the source table). \\
\midrule
\textbf{Buggy solution:}
\begin{lstlisting}[basicstyle=\ttfamily\tiny,aboveskip=2pt,belowskip=2pt]
SELECT ResourceId, AVG(Duration) AS AvgDur
FROM AccessLog
WHERE EXISTS (
  SELECT 1
  FROM UserProfile U
  WHERE U.UserId = AccessLog.UserId
    AND U.Segment = 'A'
)
GROUP BY ResourceId
\end{lstlisting}
&
\textbf{Buggy solution:}
\begin{lstlisting}[basicstyle=\ttfamily\tiny,aboveskip=2pt,belowskip=2pt]
SELECT TicketId, UpdatedAt
FROM Tickets
WHERE Status = 'Open'
ORDER BY UpdatedAt ASC
\end{lstlisting}
&
\textbf{Buggy solution:}
\begin{lstlisting}[basicstyle=\ttfamily\tiny,aboveskip=2pt,belowskip=2pt]
SELECT ProjectId, SUM(Hours) AS TotalHours
FROM ProjectMembers
GROUP BY ProjectId
\end{lstlisting} \\
\midrule
\textbf{Bug:} Opposite logic (\texttt{EXISTS} instead of \texttt{NOT EXISTS}) &
\textbf{Bug:} Wrong sort direction (should be \texttt{DESC}) &
\textbf{Bug:} Wrong source table (should use \texttt{TimeLogs}) \\
\midrule
\textbf{Error:} \textcolor{conceptred}{\textbf{WHERE}}
&
\textbf{Error:} \textcolor{conceptred}{\textbf{ORDER BY}}
&
\textbf{Error:} \textcolor{conceptred}{\textbf{FROM}} \\
\bottomrule
\end{tabular}
\end{table*}

\subsection{Evaluation on a Prerequisite-Learning Benchmark}
\label{sec:mooc-benchmark}
We evaluate \instructkg{} on the MOOC benchmark of \citet{pan2017prerequisite}, using all five released Coursera courses across two domains (Table~\ref{tab:mooc-benchmark}). We run the unmodified pipeline (Qwen2.5-14B) on the ordered transcripts and count a labeled pair as positive when the graph contains a \texttt{depends\_on} or \texttt{part\_of} edge between the two concepts. Unlike prior methods, which train supervised classifiers on the labels, \instructkg{} never observes a label, yet reaches F1 0.35 on ML and 0.29 on DSA, rising to 0.38 on ML when two hop paths are counted. Supervised classifiers reach 0.59 to 0.73 and the only training-free baseline reaches 0.29 and 0.25, which \instructkg{} surpasses on both domains. Correcting edge directions that contradict teaching order improves direction accuracy on DSA (0.62 to 0.66) but slightly decreases it on ML (0.63 to 0.62).

\subsection{Human Evaluation}
\label{sec:human-eval}
Human evaluators reviewed 100 randomly sampled concept nodes and relation triplets across all five courses. For Mechanical Engineering and Finance, the materials were annotated by the actual instructors who taught the courses; for Algorithms, NLP, and Database Systems, the annotators had subject-matter expertise in the respective domains. The evaluators judged 98.0\% of nodes and 85.0\% of triplets as correct for Algorithms, 93.0\% of nodes and 84.0\% of triplets for Database Systems, 87.0\% of nodes and 58.0\% of triplets for NLP, 86.0\% of nodes and 76.0\% of triplets for Finance, and 65.0\% of nodes and 74.0\% of triplets for Mechanical Engineering.

\paragraph{Annotator Reliability.}
To further assess reliability, two annotators independently evaluated 20 sampled nodes and 20 triplets for the NLP course. Inter-annotator agreement was substantial for both nodes (weighted $\kappa = 0.72$, 85\% exact agreement) and triplets (weighted $\kappa = 0.72$, 90\% exact agreement), with combined mean scores of $1.55/2.00$ for nodes and $1.62/2.00$ for triplets.

\subsection{Qualitative Case Study}
\label{sec:qualitative_study}
\begin{figure}[!t]
  \centering
  \includegraphics[width=0.75\columnwidth]{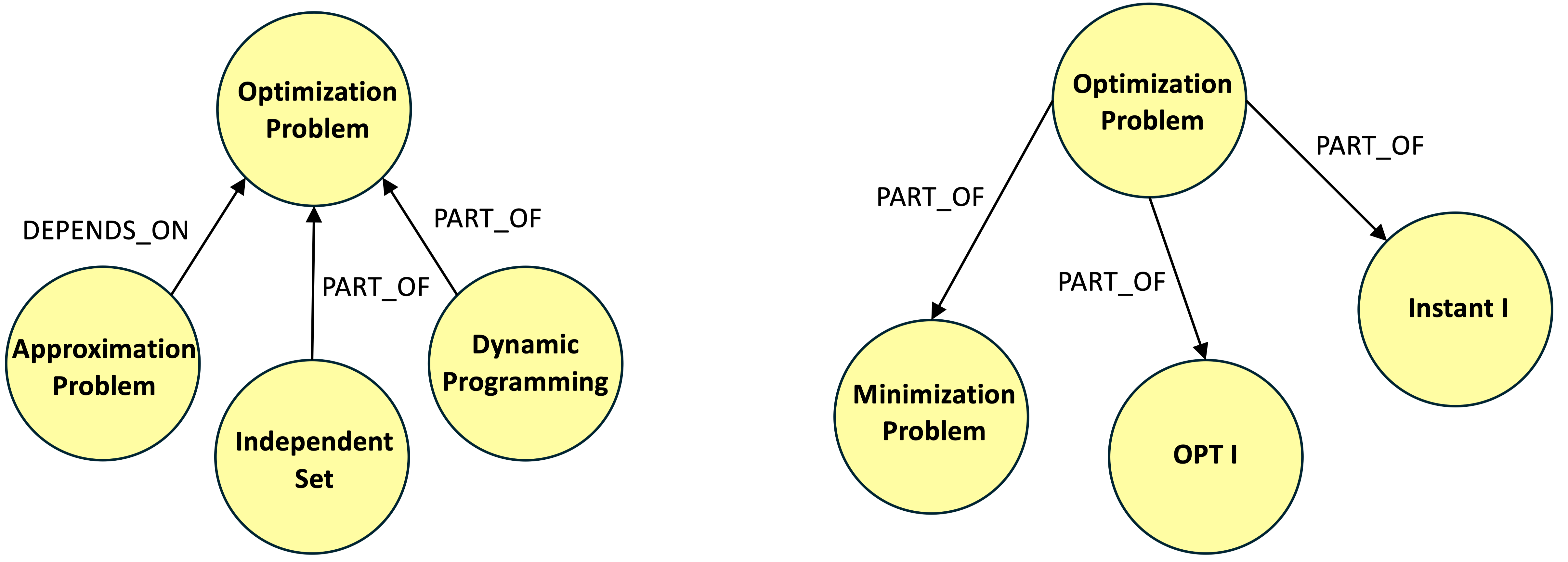}
  \caption{Comparison of knowledge graph outputs: ours (left) vs.\ KGGen (right).}
  \label{fig:main-vs-kggen_ALGO}
\end{figure}

\begin{figure}[!t]
  \centering
  \includegraphics[width=0.75\columnwidth]{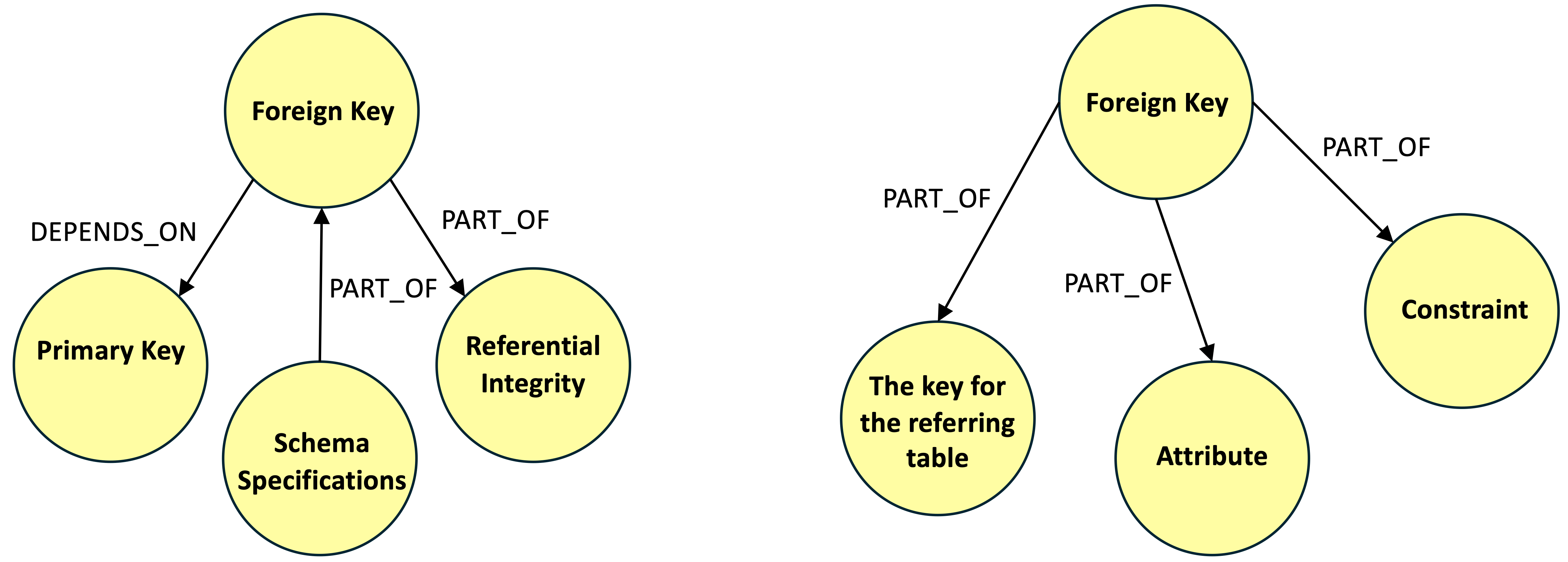}
  \caption{Comparison of knowledge graph outputs: ours (left) vs.\ KGGen (right).}
  \label{fig:main-vs-kggen_SQL}
\end{figure}

Figures~\ref{fig:main-vs-kggen_ALGO} and~\ref{fig:main-vs-kggen_SQL} compare \instructkg and KGGen subgraphs (both Qwen2.5-14B) centered on the same concept. In Figure~\ref{fig:main-vs-kggen_ALGO}, \instructkg captures that \textsc{Dynamic Programming} and \textsc{Independent Set} are part of \textsc{Optimization Problem} and that \textsc{Approximation Problem} depends on it, reflecting the course's pedagogical structure; KGGen instead links \textsc{Optimization Problem} to variable names like ``Instant I'' and ``OPT I'' that carry no pedagogical significance. This recurs across text-based methods: without structured, course-grounded concept extraction, they surface co-occurring tokens rather than meaningful concepts. In Figure~\ref{fig:main-vs-kggen_SQL}, \instructkg captures that \textsc{Foreign Key} depends on \textsc{Primary Key}, \textsc{Schema Specification} is part of \textsc{Foreign Key}, and \textsc{Foreign Key} is part of \textsc{Referential Integrity}, a coherent fragment of the prerequisite structure, whereas KGGen links \textsc{Foreign Key} to broad terms like ``attribute'' and ``constraint'' that don't reflect the intended progression.

\subsection{Student Mapping}
To illustrate a downstream application, we map student errors to specific concept gaps. Given SQL problems from a real university course and synthetic student submissions, we tag each question to graph concepts via embedding-based candidate selection and LLM concept assignment, then compare submissions against expected solutions and trace errors back to the tagged concepts. Table~\ref{tab:question-concept-mapping} shows an example from the Database Systems course: each question maps to concepts such as \textsc{Order By}, \textsc{Where}, and \textsc{From Clause}, with the error concept shown in red. A student who writes an incorrect \textsc{Where} clause in Q1 can be linked through the dependency structure to \textsc{From Clause}, a prerequisite that may also need reinforcement. Thus \instructkg's edges enable targeted diagnostics: rather than just flagging a wrong answer, the graph reveals which foundational concepts underlie the error.

\section{Discussion}
Grounding construction in instructor-provided signals yields more accurate and pedagogically meaningful graphs than LLM parametric knowledge alone. The gap is starkest at the node level, where \instructkg{} exceeds 78\% even at 3B scale while baselines remain far lower regardless of size, showing that concept quality is driven by how extraction is structured rather than by model capacity (Figures~\ref{fig:main-vs-kggen_ALGO},~\ref{fig:main-vs-kggen_SQL}). The technical contribution lies in the evidence construction preceding judgment: cluster-based aggregation roughly doubles the candidate pairs the judge can support, turning the same LLM into a stronger relation judge. The Context only comparison confirms this. Removing these signals lowers triplet accuracy on every course.

\section{Conclusion}
We presented \instructkg, a framework for constructing instructor-aligned knowledge graphs from course materials using the temporal ordering of instruction and the pedagogical roles concepts play. Across five courses it improves over two recent LLM-based methods, with node significance of 78\% or higher and triplet gains of 11 to 41 points over the strongest baseline, and traces student errors to specific concept gaps. Future work will extend the schema, add disciplines, and integrate real student assessment data.

\section*{Limitations}
Our evaluation spans five courses across computer science, finance, and engineering; broader generalization across institutions and disciplines remains an open question. The schema is restricted to two relation types (\texttt{depends\_on}, \texttt{part\_of}), and the largest model used (Qwen2.5-14B) is modest by current standards. Our human evaluation also relies on a small number of expert annotators per course, in some cases a single instructor, so the judgments may reflect individual interpretation of what constitutes a significant concept or a correct dependency.
\section*{Acknowledgments}
This work was supported in part by the Strategic Instructional Innovations Program (SIIP) and the KERN Family Foundation. We gratefully acknowledge their support.

\bibliography{references}

\appendix
\section{Baselines Details}
\label{app:baselines}
We compare \instructkg against two recent LLM-based knowledge graph construction methods, adapting each to our target schema and input format for fair comparison.

\noindent\textbf{KGGen.}
A multi-stage text-to-KG generator that extracts subject--predicate--object triples from plain text, then clusters and de-duplicates entities and edges using embedding similarity and LLM-based resolution~\cite{mo2025kggen}. KGGen extracts unconstrained predicates, which we map to our target relations (\texttt{depends\_on}, \texttt{part\_of}) using semantic similarity: we embed the extracted predicate and compare it against reference examples for each relation type (e.g., ``requires,'' ``prerequisite for'' for \texttt{depends\_on}; ``component of,'' ``subset of'' for \texttt{part\_of}), retaining only edges that exceed a similarity threshold. We apply the same chunked lecture input and enforce DAG constraints via cycle detection to ensure fair comparison.

\noindent\textbf{EDC.}
A three-phase LLM-based framework (Extract, Define, Canonicalize) that performs open triplet extraction, generates definitions for induced schema elements, and canonicalizes relations via embedding retrieval and LLM verification~\cite{zhang2024extract}. We provide EDC with a target schema containing our two relation types with definitions (e.g., ``\texttt{depends\_on}: $A$ requires $B$ as prerequisite; \texttt{part\_of}: $A$ is a component or subset of $B$'') and run the full pipeline on the same chunked lecture input. We apply global deduplication and DAG enforcement to the output for consistency.

Both baselines operate on the same lecture chunks as \instructkg but do not leverage pedagogical signals such as teaching order, concept roles, or cross-chunk thematic clustering. This comparison isolates the contribution of our instructor-aligned evidence aggregation and relation judgment approach.

\section{Prompt Templates}
\label{app:prompts}

\begin{tcolorbox}[
    colback=gray!5!white,
    colframe=gray!75!black,
    title={\textbf{Node Significance Evaluation Prompt}},
    left=2mm,
    right=2mm,
    top=2mm,
    bottom=2mm,
    breakable,
    enhanced
]

\begin{Verbatim}[fontsize=\footnotesize, breaklines=true,breakanywhere=true]
  You are evaluating a single concept node extracted for a course knowledge graph.

Course Title: {course_name}

Goal:
Decide whether the node is a SIGNIFICANT course-content concept:
something students should be taught and should understand/use in this course.

Important:
- "Significant concept" means key educational concepts (topic, method, principle, theorem,
    algorithm, framework, key technical term) for a course.
- It does NOT mean logistics/admin/metadata (e.g., instructor name, due dates, office hours,
    grading policy, LMS/Canvas, Zoom link).
- It does NOT mean educational concepts that would typically fall under a different course/topic (e.g., "Constitution", mentioned in an example of text parsing, would NOT be a significant concept for an NLP course.)

Concept Node:
- A = "{node_label}"

Use the following 0-2 ordinal scale strictly (Concept Node significance/validity):

    Definitions:
    - "Significant/meaningful concept" = a critical educational concept that students should
      be taught and be able to explain/use in this course (core topic, method, principle,
      model, theorem, algorithm, technique, framework, key term).
    - "Not meaningful for the course" includes logistical/admin/metadata items that do not
      represent course content knowledge.

    Clear examples:
    - Meaningful (YES / likely 2 if supported): "recursion", "merge sort", "Bayes' theorem",
      "photosynthesis", "supply and demand", "gradient descent", "constitutional amendments".
    - Not meaningful (NO / likely 0 even if mentioned): instructor/TA names, office hours,
      due dates, grading policy, course number, Zoom link, classroom location, required textbook ISBN,
      "homework submission", "attendance", "Canvas", "midterm".

    Scoring:
    0 = Invalid OR not a course-content concept (logistics/metadata), OR clearly insignificant or unrelated to course learning goals.
    1 = Plausible course-content concept but weakly supported, too vague, overly broad, or ambiguous given excerpts.
        (Use 1 when you cannot confirm significance from the excerpts.)
    2 = Clearly a valid course-content concept AND clearly significant for the course, with strong excerpt support.

    Evidence policy:
    - Base your score ONLY on the provided excerpts and the node label.
    - If evidence is insufficient/ambiguous, score 1 and state what is missing.
    - Cite evidence by excerpt_id (avoid long quotations).

Instructor-material excerpts (evidence base):
[formatted exerpts here]

Requirements for your answer:
- Output STRICT JSON only (no markdown).
- Rationale must cite excerpt_id(s) as evidence (e.g., "[1]", "[3]").
- If evidence is insufficient or ambiguous, score 1 and say what is missing.

Output format (STRICT JSON, no markdown, no trailing commentary):
        {
          "score": <integer in the required range>,
          "rationale": "<2-6 sentences; must reference excerpt_ids as evidence>",
          "evidence": ["<excerpt_id>", "..."],
          "notes": ["<optional brief note>", "..."]
        }
\end{Verbatim}
\end{tcolorbox}

\begin{tcolorbox}[
    colback=gray!5!white,
    colframe=gray!75!black,
    title={\textbf{Triplet Accuracy Evaluation Prompt}},
    left=2mm,
    right=2mm,
    top=2mm,
    bottom=2mm,
    breakable,
    enhanced
]

\begin{Verbatim}[fontsize=\footnotesize, breaklines=true,breakanywhere=true]
  You are evaluating a single directed typed edge (concept triplet) in a course knowledge graph.

        Task:
        Score whether the edge type and direction accurately reflect the conceptual relationship
        between the two concepts, based strictly on the instructor-provided excerpts.

        Course Title: {course_name}

        Concept Triplet:
        - A = "{edge['source']}"
        - relation = "{edge['relation_type']}"  (allowed: depends_on, part_of)
        - B = "{edge['target']}"
        Interpreting relation types:
        - depends_on: B is a prerequisite of A (B should be learned before A)
        - part_of:    B is a subtopic/component of A (A contains/organizes B)

        Use the following 0-2 ordinal scale strictly (Directed, typed edge accuracy):

    You must judge TWO things:
    (1) Whether A and B are directly related as course concepts, AND
    (2) Whether the relation TYPE AND DIRECTION are correct.

    The options for relation types are: depends_on, part_of, and None.

    If A and B should NOT be directly related, then the proposed relationship should be "None" (e.g., "mergesort", "machine learning" should not be directly related).

    Relation semantics (direction matters):
    - depends_on: Comprehending A requires understanding B; B is a prerequisite of A. Students should learn/understand B BEFORE A.
      (Read as: A depends_on B.)
    - part_of: A is a subtopic/component of B. B contains/organizes A.
      (Read as: A is part_of B.)

    Clear examples (directional):
    - Correct depends_on:
      A="merge sort" depends_on B="recursion": CORRECT (merge sort relies on recursion ideas)
      A="backpropagation" depends_on B="chain rule": CORRECT
    - Incorrect depends_on (reversed):
      A="recursion" depends_on B="merge sort": WRONG (direction reversed; recursion is more fundamental)
    
    - Correct part_of:
      A="merge sort" part_of B="sorting algorithms" and relation part_of where A part_of B: CORRECT
      A="mitochondria" with B="cell" and relation part_of where A part_of B: CORRECT
    - Incorrect part_of (reversed):
      A="sorting algorithms" part_of B="merge sort": WRONG

    Scoring:
    0 = No: the proposed direct relationship is wrong OR not supported by excerpts.
    1 = Somewhat: A and B are related, but the type and/or the direction is wrong or unclear from evidence.
    2 = Yes: A and B are directly related AND the type AND direction match the excerpts.

    Evidence policy:
    - Base your score ONLY on the provided excerpts and the proposed edge.
    - If evidence is insufficient/ambiguous, score 1 and state what is missing.
    - Cite evidence by excerpt_id (avoid long quotations).

        Instructor-material excerpts (evidence base):
        [formatted excerpts]

        Output format (STRICT JSON, no markdown, no trailing commentary):
        {
          "score": <integer in the required range>,
          "rationale": "<2-6 sentences; must reference excerpt_ids as evidence>",
          "evidence": ["<excerpt_id>", "..."],
          "notes": ["<optional brief note>", "..."]
        }
\end{Verbatim}
\end{tcolorbox}

\begin{tcolorbox}[
    colback=gray!5!white,
    colframe=gray!75!black,
    title={\textbf{Relation Judge Prompt}},
    left=2mm,
    right=2mm,
    top=2mm,
    bottom=2mm,
    breakable,
    enhanced
]

\begin{Verbatim}[fontsize=\footnotesize, breaklines=true,breakanywhere=true]

You are an expert course instructor building a concept hierarchy for a university course.

TASK:
- Use ONLY these relations: ["depends_on","part_of"].
- For this ordered pair (A,B), choose the MOST DOMINANT relation if any exists.
- **CRITICAL**: The relation direction is ALWAYS from A to B (A -> B). Do NOT reverse the direction.
- Dominance rules: prefer the single relation that best supports learning order and course understanding.
- Be minimal: avoid drawing edges just because they are possible. Only include edges that are clearly supported.
- If two relations both plausibly hold, output the dominant one only; output the other ONLY if strongly justified (lower confidence).
- You can skip making a relationship if there is no clear relation.
CAUTION: Never make a relationship between concepts that have no clear connection.

Definitions:
- Concept: a distinct course idea/skill/topic.
- Role: how a concept appears in a passage.
  - Definition = the passage defines/explains/introduces the concept.
  - Example = the passage demonstrates the concept via a concrete instance/walkthrough.
  - Assumption = the passage uses the concept as prior knowledge without teaching it.

RELATION DEFINITIONS & EXAMPLES (direction is ALWAYS A → B):

1. "depends_on": A depends_on B means A requires B as a prerequisite (B must be learned BEFORE A).
   **Direction**: A → B means "A requires B first"
   Examples:
   - "gradient descent" depends_on "derivatives" 
     -> gradient descent (A) requires derivatives (B) as prerequisite
   - "backpropagation" depends_on "chain rule" 
     -> backpropagation (A) requires chain rule (B) as prerequisite
   - "ANOVA" depends_on "variance" 
     -> ANOVA (A) requires variance (B) as prerequisite
   
   **WRONG examples (these would be backwards)**:
   - DO NOT say "derivatives" depends_on "gradient descent" (this is reversed!)
   - DO NOT say B depends_on A when A actually needs B first

2. "part_of": A part_of B means A is a component/subtype/member of the broader concept B.
   **Direction**: A -> B means "A is part of B" (A is the specific, B is the general)
   Examples:
   - "convolutional layer" part_of "neural networks" 
     -> convolutional layer (A) is a component of neural networks (B)
   - "t-test" part_of "hypothesis testing" 
     -> t-test (A) is a specific type within hypothesis testing (B)
   - "supervised learning" part_of "machine learning" 
     -> supervised learning (A) is a subcategory of machine learning (B)
   
   **WRONG examples (these would be backwards)**:
   - DO NOT say "neural networks" part_of "convolutional layer" (this is reversed!)
   - DO NOT say B part_of A when A is actually the specific case of B

PAIR:
A = "{A}"
B = "{B}"

ROLES:
{ROLE_BLOCK}

TEMPORAL / STATS:
{TEMPORAL_BLOCK}

EVIDENCE_MODE:
{MODE_BLOCK}

RULES FOR USING THE EVIDENCE TEXT: - Base your decision ONLY on the text shown in EVIDENCE below. - The EVIDENCE text is the supporting passages for this (A,B) pair. - If the text does not clearly support a relation, output null. - Your returned evidence[].quote MUST be an exact substring copied from the provided text.

EVIDENCE (use ONLY what is shown below; do not assume anything unstated):
{EVIDENCE_BLOCK}

Return strict JSON ONLY (no markdown, no extra text):
{{
  "A": "{A}",
  "B": "{B}",
  "relation": "depends_on" | "part_of" | null,
  "confidence": 0.0,
  "justification": "1-3 sentences grounded in the evidence above",
  "evidence": [
    {{"type": "chunk" | "cluster", "chunk_id": "...", "lecture_id": "...", "page_numbers": [1,2], "quote": "..."}}
  ]
}}
\end{Verbatim}
\end{tcolorbox}

\begin{tcolorbox}[
    colback=gray!5!white,
    colframe=gray!75!black,
    title={\textbf{Concept Extraction Prompt}},
    left=2mm,
    right=2mm,
    top=2mm,
    bottom=2mm,
    breakable,
    enhanced
]

\begin{Verbatim}[fontsize=\footnotesize, breaklines=true,breakanywhere=true]

You are an instructor that extracts learning concepts from text. 

- Concept: a core idea or topic about the subject matter. Only extract meaningful course concepts.
DO NOT extract example values, variable names, numbers, formulas, or code elements.
Ignore content inside examples, formulas.


Return strict JSON:
{ "concepts": ["..."] }

Rules:
- 1–5 words per concept
- No code tokens, variable names, numbers, example values
- Deduplicate
\end{Verbatim}
\end{tcolorbox}

\begin{tcolorbox}[
    colback=gray!5!white,
    colframe=gray!75!black,
    title={\textbf{Role Classification Prompt}},
    left=2mm,
    right=2mm,
    top=2mm,
    bottom=2mm,
    breakable,
    enhanced
]

\begin{Verbatim}[fontsize=\footnotesize, breaklines=true,breakanywhere=true]

You will be given a text chunk from university lecture notes or slides and a concept from the course.
Classify the role the concept plays in this chunk into one of four categories:

1**Definition**: The concept is being defined, explained, or introduced. The text describes what the concept is, its properties, or how it works.
Simple example: "Binary search is an algorithm that finds an element in a sorted array by repeatedly dividing the search interval in half."
Complex example: 
- Concept: "Big O notation"
- Text: "When analyzing algorithm efficiency, we need a formal way to describe performance. Big O notation provides an upper bound on the growth rate of an algorithm's time complexity, expressing how runtime scales with input size n."
- Classification: Definition (the concept itself is being explained, even when embedded in broader context)

2**Example**: The concept is being demonstrated or illustrated through a concrete example, walkthrough, or application. The text shows the concept in action.
Simple example: "Let's apply binary search to find 7 in [1,3,5,7,9,11]: First, check the middle element 5..."
Complex example:
- Concept: "Recursion"
- Text: "To understand how the call stack works, consider computing factorial(3). The function calls factorial(2), which calls factorial(1), which calls factorial(0) returning 1. Then factorial(1) returns 1x1=1, factorial(2) returns 2x1=2, and finally factorial(3) returns 3x2=6."
- Classification: Example (shows recursion through a concrete walkthrough, even if framed as explanation)

3**Assumption**: The concept is being used as prior knowledge, a prerequisite, or a foundation for explaining something else. The text assumes familiarity with the concept to build further understanding.
Simple example: "Using binary search, we can now efficiently implement the dictionary lookup feature..."
Complex example:
- Concept: "Hash functions"
- Text: "Hash tables achieve O(1) average-case lookup because hash functions distribute keys uniformly across buckets. However, we must handle collisions when multiple keys map to the same index."
- Classification: Assumption (hash functions are used as known background to explain hash table behavior)

4** NA **: The concept does not fall under any of the above mentioned roles: (Definition, Example, Assumption).


**Key distinction**: If the concept is being taught -> Definition. If it's being shown in action -> Example. If it's being used to explain something else -> Assumption. If none of the previously mentioned then -> NA


Also return an evidence snippet:
- Must be an exact substring copied from the chunk (10-30 words)
- Should best support your classification
- Keep it concise and relevant

Return strict JSON format only:
{ "role": "Definition" | "Example" | "Assumption" | "NA", "snippet": "..." }
\end{Verbatim}
\end{tcolorbox}

\section{Hyperparameter Details}
\label{app:hyperparams}
\begin{table}[!h]
\centering
\footnotesize
\setlength{\tabcolsep}{6pt}
\caption{Hyperparameters used across all \instructkg components.}
\label{tab:hyperparams}
\resizebox{\columnwidth}{!}{%
\begin{tabular}{llc}
\toprule
\textbf{Component} & \textbf{Parameter} & \textbf{Value} \\
\midrule
Document Ingestion
& Max tokens per chunk & 8191 \\
& Merge peers & \texttt{True} \\
\hline
\addlinespace
Concept Extraction
& Concepts per chunk & 1--5 words \\
& Deduplication & Case-insensitive \\
\hline
\addlinespace
Context Clustering
& Embedding model & all-MiniLM-L6-v2 \\
& UMAP $n_{\text{components}}$ & 15 \\
& UMAP $n_{\text{neighbors}}$ & 15 \\
& HDBSCAN $\texttt{min\_cluster\_size}$ & 5 \\
\hline
\addlinespace
Evidence Selection
& Max evidence chunks per pair & 3 \\
& Max clusters per pair & 1 \\
& Relation batch size & 8 \\
\hline
\addlinespace
LLM Inference
& Temperature & 0.1 \\
& Concurrency & 5 \\
\hline
\addlinespace
Student Mapping
& Tagging model & GPT-4o-mini \\
& Embedding model & text-embedding-3-small \\
& Candidate pool size & 60 \\
& Min confidence & 0.70 \\
\bottomrule
\end{tabular}%
}
\end{table}

\section{Confidence Filtering Analysis}
\label{app:filter}
\begin{table}[!h]
\centering
\footnotesize
\caption{Triplet accuracy under evidence based filters (Qwen2.5-14B, all five courses).}
\label{tab:filter-curve}
\begin{tabular}{lcc}
\toprule
\textbf{Filter} & \textbf{Edges kept} & \textbf{Accuracy} \\
\midrule
All edges & 100\% & 0.567 \\
2+ evidence chunks & 51\% & 0.562 \\
3+ evidence chunks & 11\% & 0.544 \\
Co-occurrence evidence only & 37\% & 0.536 \\
Cluster evidence only & 25\% & 0.556 \\
\bottomrule
\end{tabular}
\end{table}

\noindent No evidence based filter improves accuracy over the unfiltered graph, so we retain all edges and expose per edge evidence instead.

\section{Runtime Breakdown}
\label{app:runtime}
For the representative course (Qwen2.5-14B, single GPU): extraction 6m12s (1{,}334 LLM calls), clustering and pair construction 21s (no LLM calls), relation judgment 48m03s (3{,}373 calls), total 54m36s. Relation judgment accounts for 85 to 90\% of wall clock time across all corpora we processed; totals range from 3 minutes for a small slide course to 3.6 hours for the largest transcript collection.

\end{document}